\documentclass[runningheads]{llncs}
\usepackage{graphicx}
\usepackage{amsmath,amsfonts,amssymb}
\usepackage{hyperref}
\usepackage{csquotes}
\usepackage{cite}
\usepackage{multirow}
\usepackage[export]{adjustbox}

\begin{document}
\title{A Framework for Explainable Concept Drift Detection in Process Mining}
%
%
\author{Jan Niklas Adams\inst{1}\and
Sebastiaan J. van Zelst\inst{2,1}\and
Lara Quack\inst{2} \and
Kathrin Hausmann\inst{2} \and
Wil M.P. van der Aalst\inst{1,2}\and
Thomas Rose\inst{2,1}}
\authorrunning{J. N. Adams et al.}
%
\institute{RWTH Aachen University, Germany \\
\email{\{niklas.adams, s.j.v.zelst, wvdaalst\}@pads.rwth-aachen.de}\\ \and
Fraunhofer Institute for Applied Information Technology (FIT), Germany\\
\email{\{lara.quack, kathrin.hausmann, thomas.rose\}@fit.fraunhofer.de}}
\maketitle              
\begin{abstract}
Rapidly changing business environments expose companies to high levels of uncertainty. 
This uncertainty manifests itself in significant changes that tend to occur over the lifetime of a process and possibly affect its performance. 
It is important to understand the root causes of such changes since this allows us to react to change or anticipate future changes. Research in process mining has so far only focused on detecting, locating and characterizing significant changes in a process and not on finding root causes of such changes.
In this paper, we aim to close this gap.
We propose a framework that adds an explainability level onto concept drift detection in process mining and provides insights into the cause-effect relationships behind significant changes. We define different perspectives of a process, detect concept drifts in these perspectives and plug the perspectives into a causality check that determines whether these concept drifts can be causal to each other.  We showcase the effectiveness of our framework by evaluating it on both synthetic and real event data. Our experiments show that our approach unravels cause-effect relationships and provides novel insights into executed processes.

\end{abstract}
\section{Introduction}
Digitization poses great threats but also exceptional opportunities to companies. On the one hand, new technologies, business models, and legislation expose companies to high levels of uncertainty. On the other hand, the introduction of information systems over the last decades enables companies to collect and analyze data on their \textit{business processes}. These data can be converted into an \textit{event log} and are used to discover, monitor and improve the underlying business processes. It, thus, helps the companies to deal with the uncertainty they are exposed to. \textit{Process mining} \cite{ProcessMiningDSIA} is the discipline of computer science that successfully analyzes and improves business processes by 
applying concepts of process and data science to transform event logs into process models and actionable insight for the process owner.
When looking at business processes uncertainty is often caused by significant change, called \textit{concept drift}, in some perspective of a business process. For example, due to Covid-19 a lot of companies had to redesign or extend processes by  including digital alternatives to previously in-person activities. This resulted in so-called concept drifts. As the quality and profitability of organizations highly depend on their business processes concept drift can have a huge impact on either of these dimensions. The restructuring of a process to meet the health safety regulatory standards could, e.g., lead to increased processing time and thus increased cost. Detecting and handling concept drift has, thus, been named one of the main challenges in process mining \cite{manifesto}. For the process owner the mere knowledge of past occurrences of concept drifts is not sufficient. To derive useful insights it is helpful to know the underlying cause-effect relationships associated with these concept drifts, i.e., the user can either anticipate future concept drifts or use the uncovered relations to further improve the process.

\begin{figure}[t]
\centering
\includegraphics[width=1\textwidth]{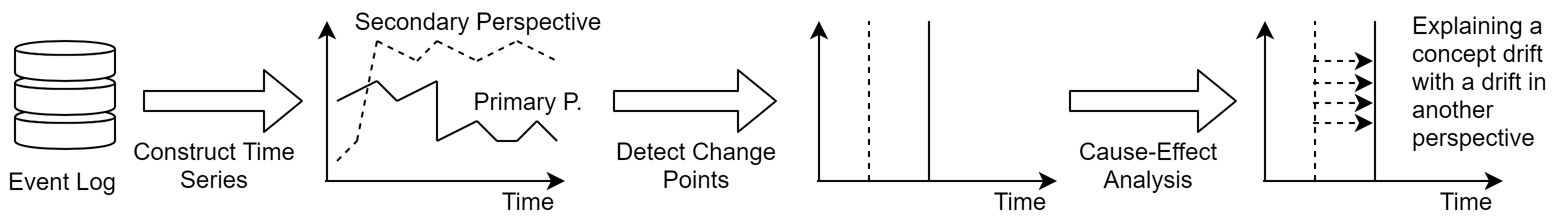}
\caption{Our proposed framework transforms event data into two time series representations that can describe various different process perspectives, e.g., the weekly workload. A cause-effect analysis is then conducted using the information of detected concept drifts in the selected perspectives to unravel root causes for these drifts. } \label{fig:framework_simple}
\end{figure}

In this paper, we introduce a generic framework that augments concept drift detection in process mining by adding a cause-effect analysis on top of the detected concept drifts. This cause-effect analysis extracts possible explanations for the occurrence of a concept drift. 
The core idea of the framework is depicted in \autoref{fig:framework_simple}. 
Before starting, we choose a perspective of the process to be analyzed for concept drift, i.e., the \textit{control-flow}, \textit{data}, \textit{resource} or \textit{performance} perspective. This perspective is called the \textit{primary} perspective. As we are interested in the root causes for these concept drifts, we choose a \textit{secondary} perspective that could contain root causes. This secondary perspective is also analyzed for concept drifts and these are tested for causality with the concept drifts of the primary perspective.
In the first step, we transform an event log into two time series for both the chosen primary and secondary perspective. After detecting concept drifts in both perspectives, we conduct a cause-effect analysis and check which concept drifts of the secondary perspective could be causal to a concept drift in the primary perspective. The set of explainable concept drifts forms the output of our framework.

Our framework touches the areas of concept drift detection and cause-effect analysis in process mining which have, thus far, primarily been studied separately. Most of the work on concept drift deals with detecting concept drifts and only considers the control-flow perspective of a process. The control-flow perspective describes the structuring and dependencies of activities. Recent work, e.g., the approach of Brockhoff et al. \cite{CPDEarthMoversDistance} introduces additional perspectives, i.e., the time perspective to concept drift detection. Ostovar et al. \cite{CPDAlphaPlusRelations} add an additional characterization of the drift, i.e., providing information about underlying nature of the drift, on top of the mere detection of the drift.
With our work, we include more perspectives and add an explainability level to concept drift detection. 
Most work on cause-effect analysis uncovers cause-effect relationships on a process-instance-level, e.g., giving recommendations for individual customers to maximize the outcome, as Bozorgi et al. \cite{RCAcausalmloutcome} recently introduced. Pourbafrani et al. \cite{mahsaSASD} focus on finding cause-effect relationships on a global-process level and use these to simulate what-if scenarios. 

The remainder of this paper is structured as follows. We introduce the related work on concept drift detection and cause-effect analysis in \autoref{sec:related_work}. In \autoref{sec:preliminaries}, we provide the definitions and background used in the remainder of this paper. We illustrate our framework for explainable concept drift detection in \autoref{sec:framework}. In \autoref{sec:evaluation}, we provide details on our specific implementation and evaluate our framework with synthetic data and conduct a case study on real-life event data. \autoref{sec:conclusion} concludes the paper.

\section{Related Work}
\label{sec:related_work}
A general introduction to the field of process mining is given in \cite{ProcessMiningDSIA}. In this section, we introduce related work on concept drift detection and cause-effect analysis in process mining.

\textit{Concept drift detection} (also: change point detection), has received much attention outside of process mining. A general introduction can be found in \cite{SurveyCPD}. As our use case does not provide labeled data sets for supervised algorithms, we are only interested in unsupervised concept drift detection algorithms as the training data for supervised algorithms is expensive to obtain and the ground truth is hard to define in the setting of real-life event logs.
Existing work on concept drift detection in process mining focuses on the detection, localization and characterization of changes, not the explanation of them.
Detection refers to the presence of a concept drift, localization to the time of occurrence of the drift and characterization to the nature of the drift, e.g., whether an activity was removed or a performance indicator significantly increased. Explanation refers to the root causes of a drift, e.g., why an activity was removed or why a performance indicator significantly increased.
Most of the work aims to detect drifts in the control-flow perspective. Bose et al. \cite{DealingDriftPM, HandlingDriftPM} and Martjushev et al. \cite{CPDGradualMultiOrder} built representations of the control-flow perspective by using the (directly) follows relations. They employ hypothesis testing to compare a window of values before and after a potential change point for significant differences. Maaradji et al. \cite{suddengradualdriftostovar} and Ostovar et al. \cite{CPDAlphaPlusRelations} use the $\alpha$ and $\alpha^+$-relations \cite{AlphaPlus} to model the control-flow perspective while also using hypothesis testing to determine change points. 
\bgroup
\setlength{\tabcolsep}{0.5em}
\begin{table}[t]
    \centering
    \caption{Overview of the related approaches for cause-effect analysis and concept drift detection in process mining. Cause-effect analysis can be performed either on the case- or the process level. Approaches for concept drift detection have different scopes, i.e., they can detect, locate, characterize or explain a concept drift. Our approach detects, locates and explains concept drift and therefore yields insights into process level cause-effects.}
     \resizebox{0.95\textwidth}{!}{
    \begin{tabular}{|l|c|c|c|c||c|c|c|}\hline
         &\multicolumn{4}{c||}{Concept Drift}  &\multicolumn{2}{c|}{Cause-Effect Analysis} \\ \hline
         &{Detect.}&{Locat.}&{Char.}&{Expl.}&{case level}&{process level}\\\hline
         \cite{HandlingDriftPM,DealingDriftPM,CPDEarthMoversDistance,CPDGradualMultiOrder,CPDHypothesisTestingAdaptiveWindow,CPDWorkflowChangesDistanceMatrix,CPDOnlineWeber,CPDParikhVectors,TsinghuaProcessMining,suddengradualdriftostovar} &\checkmark&\checkmark&&&&\\\hline
         \cite{CPDAlphaPlusRelations,CPDTraceClustering,graphbaseddetection,ProcessDriftVisualAnalytics} &{\checkmark}&{\checkmark}&{\checkmark}&&& \\\hline
         \cite{RCAgenericframeworkdecisiontree,RCAcontextawareperformanceanalysis,RCArootcauseenriched,RCADataAwarePrediciton,RCAcausalmloutcome,RCAgrangercausalfactors}&&&&&{\checkmark}&  \\\hline
         \cite{mahsaSBP,mahsaSASD,PMSD,analyzingresourceBehaviourUsingProcessMining}&&&&&&{\checkmark} \\\hline
         
          \hline
         Our approach &{\checkmark}&{\checkmark}&&{\checkmark}&&{\checkmark}\\\hline
    \end{tabular}
    }
    \label{tab:related_work}
\end{table}
\egroup
One notable recent approach is the one of Yeshchenko et al. \cite{ProcessDriftVisualAnalytics}. The authors use \textsc{DECLARE} constraints to model the control-flow of a process. They define a range of time windows and subsequently calculate values for the declarative constraints for each time window, forming a multivariate time series. This time series is analyzed for concept drifts by applying the Pruned Exact Linear Time algorithm. By visualizing the results of this clustering, the user can characterize the occurring concept drifts.
Other authors also include other perspectives than control-flow. Leontjeva et al. \cite{PPMSequenceEncodingDataPayload} and Meisenbacher et al. \cite{PPMHandlingConceptDrift} use the data payload of past events to include the data perspective into their representation.
Analyzing the related work on concept drift in process mining reveals two shortcomings: The little consideration of additional perspectives other than control-flow and the absence of root cause analysis for concept drifts. With this paper we aim to close this gap.

The area of cause-effect analysis in process mining investigates relationships that are present in a process.
One way to define different levels of analysis is to either consider the local intra-trace, i.e., case level, or the global level of the process. The case level deals with individual process instantiations, e.g., a customer running through the process of applying for a loan. The global level is the entirety of components and cases that are associated with the process. Many approaches in cause-effect analysis focus on the case level rather than the global level, providing recommendations and predictions for handling individual cases. De Leoni et al.  \cite{RCAgenericframeworkdecisiontree} and Hompes et al. \cite{RCAcontextawareperformanceanalysis} provide methods to extract root causes for performance variations on a case level. 
In another work, Hompes et al. \cite{RCAgrangercausalfactors} group events based on certain process performance characteristics and further decompose these groups based on different characteristics. They subsequently test for cause-effects between these characteristics by looking at their development over time and testing for Granger-causality \cite{GRANGERcausality} to extract the root causes of performance variations on a case level. This technique works well for identifying causal factors for performance variations, however, other perspectives of the process such as control-flow or resources are, so far, not considered, potentially missing important cause-effects.
Bozorgi et al. \cite{RCAcausalmloutcome} use techniques from causal machine learning to provide recommendations for handling an individual case that maximize the probability of a certain outcome. These approaches provide useful information for individual cases, however, they are not able to detect important cause-effect relationships, that happen on a global level, e.g., an increase in customers, that leads to longer waiting times.
Other authors investigate cause-effect relationships on a global level. Pourbafrani et al. \cite{mahsaSBP,mahsaSASD} use system dynamics as a modeling tool of the process over time
and construct a model that contains cause-effect relationships between different metrics. This model is subsequently used to simulate the outcomes for different scenarios. Nakatumba et al. \cite{analyzingresourceBehaviourUsingProcessMining} investigate the effect of resource workload on their performance using regression analysis.

A selection of papers on concept drift and cause-effect analysis in process mining is depicted in \autoref{tab:related_work}.
Our framework is the only technique that covers both spectra.

\section{Preliminaries}
\label{sec:preliminaries}
In this section, we introduce the core definitions of this paper and the main notations used for improving the readability.
$\mathcal{P}(X){=}\{X^{\prime} {\mid} X^\prime {\subseteq} X\}$ denotes the power set of a set $X$. A sequence allows enumerating the elements of a set. A sequence of length $n$ over $X$ is a function $\sigma {:} \{1,\ldots,n\} {\rightarrow} X$ which we write as $\sigma {=} \langle \sigma (1), \sigma (2), \ldots, \sigma(n) \rangle$.

An event can be considered the \enquote{atomic datum} of process mining. An event consists of values that are assigned to attributes, e.g., the executed activity, the timestamp, a case-id and other attributes. Each line in \autoref{tab:log} corresponds to an event. Each event needs to be assigned a case-id describing the process instance which is the \textit{case} this event belongs to. All lines with the same case-id in \autoref{tab:log} form a case. The collection of recorded cases forms an event log.
\begin{table}[t]
    \centering
     \caption{Exemplary event log with two cases and resources handling the activities}
     \resizebox{0.48\textwidth}{!}{
    \begin{tabular}{|c|c|c|c|}
    \hline
         case-id&{activity}&{timestamp}&{resource} \\\hline
         1&{register}&{2021-06-15 12:30}&{Peter}\\\hline
         1&{submit}&{2021-06-15 12:35}&{Sophia}\\\hline
         2&{register}&{2021-06-15 13:12}&{Peter}\\\hline
         1&{reply}&{2021-06-15 14:21}&{Christina}\\\hline
    \end{tabular}
   }
    \label{tab:log}
\end{table}
\renewcommand{\labelitemi}{\textbullet}
\begin{definition}[Events, Cases and Projections]
\label{def:def1}
An event describes the information associated to the execution of an activity. Let $\mathcal{E}$ denote the universe of events. Let $\mathcal{D}$ denote the universe of attributes and let $\mathcal{V}$ denote the universe of possible attribute values. Let $\mathcal{T}$ denote the universe of possible timestamps.
\begin{itemize}
    \item For an attribute $d\in \mathcal{D}$ we assume the existence of a mapping to retrieve the corresponding attribute value $\pi_d^\mathcal{E}{:}\mathcal{E}{\nrightarrow}\mathcal{V}$.
    \item The activity projection is a total function retrieving the activity of an event $\pi_{act}^\mathcal{E}:\mathcal{E}\rightarrow \mathcal{A}$, where $\mathcal{A}$ denotes the universe of activities.
    \item The time projection is a total function retrieving the timestamp of an event $\pi_{time}^\mathcal{E}{:}\mathcal{E}{\rightarrow}\mathcal{T}$.
    \item Each event has an identifier $\pi_{id}^{\mathcal{E}}$ to differentiate between events, that might have the same values for each attribute. Therefore, $e,e^\prime{\in}\mathcal{E}(\pi_{id}^\mathcal{E}(e){=}\pi_{id}^\mathcal{E}(e^\prime) \Rightarrow e{=}e^\prime)$
    
\end{itemize}
Events belong to a case denoting the process instance of this event. Let $\mathcal{C}$ be the universe of all cases. 
\begin{itemize}
    \item For an attribute $d{\in}\mathcal{D}$ we assume the existence of a projection function to retrieve the corresponding attribute value $\pi_d^\mathcal{C}:\mathcal{C}{\nrightarrow}\mathcal{V}.$
    \item Each event $e{\in}\mathcal{E}$ has a case-id describing the case it belongs to. The projection function $\pi_{case}^\mathcal{E}(e)$ retrieves the corresponding case-id.
    \item We furthermore assume the existence of an event projection that maps a case on the set of its events $\pi_{events}^\mathcal{C}{:} \mathcal{C}{\rightarrow}\mathcal{P}(\mathcal{E})$.Cases are non-overlapping, i.e., $\forall c_1,c_2{\in}\mathcal{C}(\pi_{events}^\mathcal{C}(c_1) {\cap} \pi_{events}^\mathcal{C}(c_2){\neq} \emptyset \Rightarrow c_1{=}c_2)$.
\end{itemize}
 An event log $L$ is a set of cases, thus $L{\in}\mathcal{P}(\mathcal{C})$. 
 \begin{itemize}
     \item For an attribute $d{\in}\mathcal{D}$ we assume the existence of the projection functions $\pi_d^\mathcal{E}(L) {=} \{\pi_d^\mathcal{E}(e) {\mid} \exists_{c{\in} L} e {\in} \pi_{events}(c)\}$ and $\pi_d^\mathcal{C}(L){=} \{\pi_d^\mathcal{C}(c) {\mid} c {\in} L\}$ to retrieve the set of values for this attribute.
 \end{itemize}
\end{definition}
\section{Generic Framework for Explainable Concept Drift}
\label{sec:framework}
\begin{figure}[t]
\centering
\includegraphics[width=0.93\textwidth]{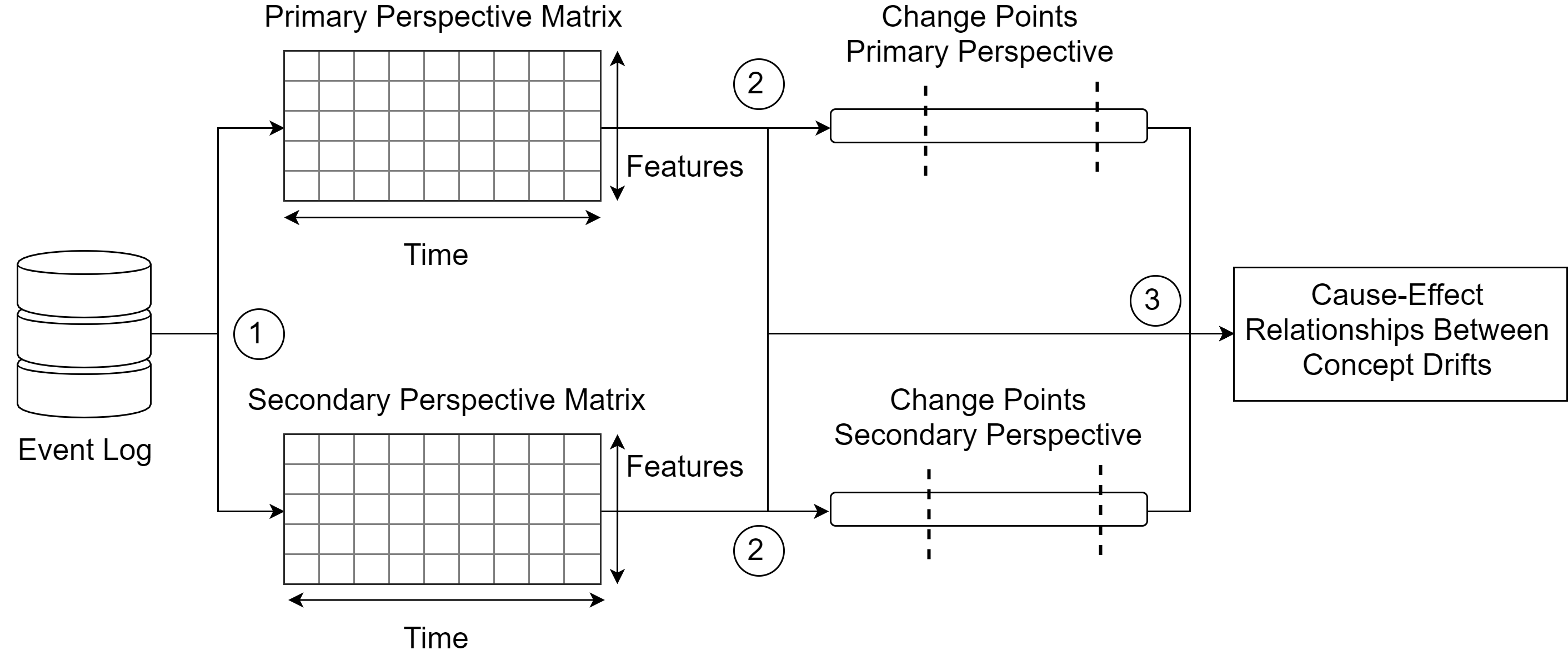}
\caption{Our proposed framework for uncovering cause-effect relationships. In Step 1, the event log is transformed into two time series representations of different perspectives. In Step 2, the change points are detected and checked for causality in Step 3.} \label{fig:framework}
\end{figure}

In this section, we introduce the three steps of our framework. 
The framework is depicted in \autoref{fig:framework}. In the first step, we construct time series representations for both chosen perspectives. In the second step, a change point detection algorithm is performed on both perspectives. In the third step, we test pairs of change points for causality by taking the time lag between them and check whether the two time series can be causal given the lag.  If this test is positive the detected concept drift together with the cause-effect-relationship are as explanation given to the user. For each of the steps, we provide the input and output specifications.
\subsection{Time Series Construction}
To express the development of a process perspective over time, we construct a time series. A time series assigns values to subsequent time intervals. We, therefore. need to map an event log onto time intervals and then assign values to these intervals. For assigning values to a set of events that are contained in a specific time interval, we first need a way to look at these events in isolation and thus define a selection function for events based on a time interval.
\begin{definition}[Time Intervals]
Based on a reference timestamp $t_r{\in}\mathcal{T}$, e.g., describing the beginning of an event log, we can express every timestamp as a real number that describes the time, e.g., number of milliseconds or nanoseconds, passed since this reference timestamp. The real-valued representation of a timestamp can be retrieved with the function $r_{t_r}(t){\in}\mathbb{R}$. 
Given two timestamps $t_1, t_2 {\in} \mathcal{T}$ and $t_1^\mathbb{R}{=}r_{t_r}(t_1), t_2^\mathbb{R}{=}r_{t_r}(t_2)$ we define a time interval $ti{=} [t_1^\mathbb{R},t_2^\mathbb{R})$ where $t_1^\mathbb{R} {<}  t_2^\mathbb{R}$. Let $\mathcal{TI}$ denote the universe of possible time intervals. ${|}ti{|} = t_2^\mathbb{R} {-} t_1^\mathbb{R}$ defines the length of a time interval and $start(ti)=t_1^\mathbb{R}$. To extract the events of an event log $L$ occurring within a time interval $ti {\in} \mathcal{TI}$ we define the selection function $sel(L,ti){=}\{e{\in}\mathcal{E} {\mid} \exists_{c{\in}L}e{\in}\pi_{events}^\mathcal{C}(c) \wedge r_{t_r}(\pi_{time}^\mathcal{E}(e)){\in}ti\}$. 
\end{definition}
This definition is used to map an event log onto time intervals. To finish constructing the time series we now calculate a value for each time interval. We first need to define a function to map a collection of events onto a real-valued number and can then extend this to multiple functions, mapping a collection of events onto a real-valued vector and finally a time series, doing this for multiple subsequent time intervals.
\begin{definition}[Event Log to Time Series]
\label{def:feature}
Given a time interval $ti{\in}\mathcal{TI}$ and an event log $L$, we define the function $f(sel(L,ti)) {\in} \mathbb{R}$ that maps an event log for a specific time interval onto a real-valued number. We use the notation $f(L,ti){=}f(sel(L,ti))$ for readability. Let $f_1,f_2,\ldots,f_m$ be functions of the signature of $f$. We define the function $g(L,ti)=(f_1(L,ti),f_2(L,ti),\ldots,f_m(L,ti))$ with $g(L,ti){\in}\mathbb{R}^m$ to construct a real-valued vector for a specific time interval of an event log. Let $TI{\in}\mathcal{TI}^*$ define a mutually exclusive sequence of time intervals of equal lengths, i.e., $\forall ti_i,ti_j{\in}TI ({|}ti_i{|}{=}{|}ti_j{|} \wedge i{\neq}j {\Rightarrow}  ti_i{\cap}ti_j{=}\emptyset \wedge i {<}j \Rightarrow start(ti_i){<}start(ti_j))$. With $TI{=}\langle ti_1,\ldots, ti_n\rangle$ we define the time series construction function $h_{g,TI,L}{=}(g(L,ti_1)^\intercal,\ldots,g(L,ti_n)^\intercal)$ with $h_{g,TI,L}{\in}\mathbb{R}^{m\times n}$ to retrieve a real-valued matrix, that represents a multivariate time series of an event log.
\end{definition}

For each perspective there are many ways to represent it as a real-valued number. Take the control-flow as an example. We can count the number of distinct activities for subsequent time intervals. If the number of activities suddenly increases we know that there is a concept drift in the control-flow since a new activity was added. Different measurements can be combined as a single one can not represent the whole perspective. E.g., if one activity was removed and one was added taking the number of distinct activities as representation could not express this.
\autoref{tab:feature_representations} depicts a non-exhaustive list of measurements for each of the introduced perspectives. For a detailed introduction we refer to the corresponding papers.

For the control-flow perspective, we can use simple measures as, e.g., the number of distinct activities or intermediate results of mining algorithms like the $\alpha$-relations obtained from the $\alpha$-Miner \cite{alpha}. 

For the performance perspective we can leverage heavily on the recorded times, which can be seen as a proxy for cost or service quality. We can, e.g., calculate the average service times for each activity, i.e., the time from start to completion of an activity. Furthermore, we can define a threshold of processing time and count all the cases that exceed this threshold and are thus classified as overtime.

The measures for the data perspective use the additional attributes associated to events, e.g., the age or credit score of customers. We can use aggregation functions such as average or maximum to map all the values of an attribute in a time interval onto a single number, e.g., the average age of customers for each time interval. We can, furthermore, count the number of events to describe the event volume over time.

Representations for the resource perspective rely on information about the resources, often staff members, handling an activity. We can count the number of events a resource is involved in to calculate the workload and its development over time. By simply counting the number of active resources each time frame, we can furthermore monitor the number of deployed resources over time.

The question remains which perspectives a user should choose. There is not a general answer for this, domain knowledge and potential assumptions can be used. However, the investigation of certain perspectives might be more promising than others. There are some examples of reoccurring cause-effect themes in process mining. Resources often have an impact onto the performance of a process, e.g., the workload onto the service times \cite{analyzingresourceBehaviourUsingProcessMining}, the workload onto overtime cases \cite{RCArootcauseenriched} or the associated data onto the case duration \cite{RCADataAwarePrediciton}. Furthermore, e.g., control-flow changes such as changing prevalence of a choice might influence the performance perspective. 
\setlength{\tabcolsep}{0.3em} 
\begin{table}[t]
    \centering
    \caption{Possible mapping functions to construct a real-valued representation of different perspectives of a business process.}
    \resizebox{0.96\textwidth}{!}{
    \begin{tabular}{|l|l|l|l|}\hline
    \multicolumn{1}{|c|}{Control-Flow}&\multicolumn{1}{c|}{Performance}&\multicolumn{1}{c|}{Data}&\multicolumn{1}{c|}{Resources}\\\hline
          Directly-follows & Service times\cite{analyzingresourceBehaviourUsingProcessMining}&Aggregation of&Workload\cite{analyzingresourceBehaviourUsingProcessMining}\\
          frequencies \cite{alpha}&&case attributes&\\\hline
          $\alpha$-relations \cite{alpha}&Overtime cases \cite{RCArootcauseenriched}&Aggregation of& Involved resources\\
          &&activity attributes&\\\hline
          $\alpha^+$-relations \cite{AlphaPlus}&Case durations\cite{RCAgrangercausalfactors}&Number of&Number of \\
          &&events or cases&active resources\\\hline
          Heuristic Miner's&Activity sojourn&Threshold&Aggregation of \\
          $a \Rightarrow_W b$ score \cite{HeuristicsMiner}&time\cite{RCAgrangercausalfactors}&exceedings& attribute values\\\hline
          Number of activities&Activity waiting time \cite{RCAgrangercausalfactors}&&\\\hline
          
          \textsc{DECLARE} constraints \cite{ProcessDriftVisualAnalytics}&&&\\ \hline
          $\ldots$&$\ldots$&$\ldots$&$\ldots$\\\hline
    \end{tabular}
    }
    \label{tab:feature_representations}
\end{table}
\subsection{Change Point Detection}
After constructing multivariate time series for each the primary and the secondary perspective, we detect change points in these time series. The change point detection technique maps a time series onto subsets of the time intervals in which the distribution of the features significantly changed. 
\begin{definition}[Change Point Detection in Multivariate Time Series]
Let $H{\in} \mathbb{R}^{m\times n}$ be a multivariate time series and $TI{=}\langle ti_1,\ldots,ti_n\rangle$ be the previously introduced sequence of time intervals used to construct this time series. A change point detection technique $CPD$ maps a time series onto a subset of time intervals, where a significant change of the underlying time series occurred $CPD(H) {\subseteq} TI$.
\end{definition}
As an input the change point detection method has to be able to process a multivariate time series. As mentioned in \autoref{sec:related_work}, this method should be unsupervised, i.e., be able to detect change points without seeing similar kinds of time series with annotated change points before.
Examples of change point detection techniques include cost-based techniques \cite{SurveyCPDCostFunction}, hypothesis testing \cite{DealingDriftPM} or clustering techniques \cite{CPDTraceClustering}.

\subsection{Cause-Effect Analysis}
In the first step, two time series for the primary and the secondary perspective are constructed. A set of $m_p$ mapping functions for the primary perspective and $m_s$ for the secondary perspective are applied to construct the time series $H_p{\in}\mathbb{R}^{m_p{\times} n}$ and $H_s{\in}\mathbb{R}^{m_s {\times} n}$.
The change point detection step of the framework calculates two sets of change points $CPD(H_p)$ and $CPD(H_s)$. In the cause-effect analysis step, we look at the change points in the primary perspective and analyse which concept drifts in the secondary perspective potentially have a cause-effect relationship to a concept drift in the primary perspective. We, therefore, look at each primary drift and consider all preceding secondary drifts. We calculate the time lag, i.e., the number of time intervals that lie between the drifts, and test whether the secondary perspective can potentially be causal to the primary perspective given this lag.  We, therefore, test all pairs of features between primary and secondary perspective for causality given this lag. A feature is a row of the time series describing a single measurement over time. If a feature pair is tested to be causal we add the change point pair and all causal feature pairs to the output of our framework. 
\begin{definition}[Cause-Effect Analysis]
Let $H_p{\in}\mathbb{R}^{m_p{\times} n}$ and $H_s{\in}\mathbb{R}^{m_s {\times} n}$ be time series for the primary and secondary perspective. 
We define the lag function $l_{TI}(ti_1,ti_2){\in}\mathbb{N}$ to retrieve the number of time intervals in $TI$ that lie in between $ti_1$ and $ti_2$, i.e., the lag. For a change point of the primary perspective $cp_p{\in}CPD(H_p)$ and a change point in the secondary perspective $cp_s{\in}CPD(H_s)$ we retrieve the lag $k$ using the lag function $k=l_{TI}(cp_p,cp_s)$. Given a row, i.e., a feature, of the primary and secondary perspective $h_p{\in}H_{p,i}, i{\in}\{1,\ldots,m_p\}$ and $ h_s{\in}H_{s,i}, i{\in}\{1,\ldots,m_s\}$, where $h_s,h_p{\in}\mathbb{R}^n$, a cause-effect analysis technique $CA$ maps these two features and a time lag $k$ onto a value between 0 and 1 
$CA(h_p,h_s,k){\in} [0,1]$.
This value indicates whether a cause-effect relationship with lag $k$ is present or not.
\end{definition}
The set of all change point pairs with all detected cause-effect relationships between feature pairs forms the output of the framework.
\section{Evaluation}
\label{sec:evaluation}
\subsection{Implementation}
\label{sec:implementation}
We implemented our framework on the basis of PM4Py \cite{pm4py}. The implemented version is available at GitHub\footnote{\url{https://github.com/niklasadams/explainable_concept_drift_pm.git}}. In this section, we introduce the techniques in change point detection and cause-effect analysis specific to our implementation.

Similar to Yeshchenko et al. \cite{ProcessDriftVisualAnalytics}, we use the Pruned Exact Linear Time PELT-algorithm \cite{PELTdescription} as a change point detection technique for multivariate time series. This technique uncovers change points by minimizing a cost function that depends on assigning change points.
It is able to process a multivariate time series and computes an optimal solution in linear time and is therefore well suited for our experiments. An exact description can be found in \cite{PELTdescription}. For applying the PELT-algorithm a penalty $\beta$ has to be chosen that prevents overfitting. The calculated change points are subsequently used to calculate the lags needed for cause-effect analysis. We use the concept of Granger-causality \cite{GRANGERcausality}. Granger-causality determines with which probability two time series are correlated given a time lag between them and can thus be seen as a type of predictive causality. The user has to provide a p-value that describes the threshold probability at which feature pairs should be classified as Granger-causal.

\subsection{Experiments}
\begin{figure}[t]
\centering
\includegraphics[width=0.99\textwidth]{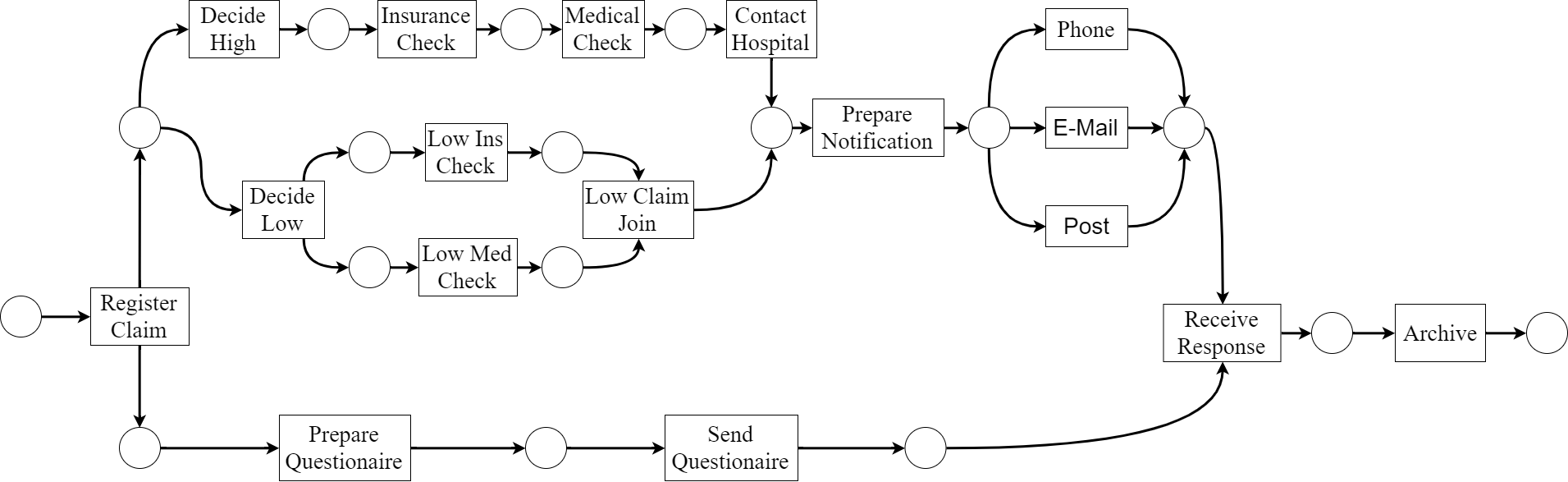}
\caption{Synthetic process for claiming insurance. The availability of different ways of notification, phone, e-mail and post, depends on the age of the customer. \cite{HandlingDriftPM}} \label{fig:synthetic_process}
\end{figure}
We evaluate our framework using a synthetic event log and a real-life event log. The synthetic log is used as a means to verify the implementation and a proof of concept. We, then, expand this to conduct a case study on real-life event data and discuss our findings. For both experiments we provide the chosen parameters for the three steps of our framework, i.e., perspectives and measurements in time series construction, change point detection in multivariate time series and cause-effect analysis. To verify our results, we compare our findings to state-of-the-art methods in concept drift detection and cause-effect analysis.
\subsection{Synthetic Insurance Event Log}

We use CPN Tools \cite{CPNTools} to generate an event log based on a simulation model of claiming insurance shown in \autoref{fig:synthetic_process}. The different ways of notification, i.e. postal, phone and email, are available for different ages of customers. When simulating we introduce a drift in the ages of customers that should then cause a succeeding drift in the prevalence of notification activities, especially an increase in email-notification. 
The chosen parameters for the instantiation of our framework are depicted in \autoref{tab:par_synth}. 
\setlength{\tabcolsep}{0.35em} 
{\renewcommand{\arraystretch}{1.2}
\begin{table}[t]
    \centering
    \caption{Parameter choices for running the framework on the synthetic event log.}
    \resizebox{0.85\textwidth}{!}{
    \begin{tabular}{|l||l|l|l|}\hline
    \multicolumn{4}{|c|}{Experimental Setup}\\\hline
         \multirow{4}{*}{\shortstack[l]{Time series\\ construction}}&Primary perspective& Control-Flow& Directly-follows frequencies\\\cline{2-4}
         &\multirow{2}{*}{Secondary perspective}& \multirow{2}{*}{Data} & Minimum, maximum, sum,\\
         &&& average, count, set average\\\cline{2-4}
         &Time interval duration & \multicolumn{2}{c|}{1 day}\\\hline
          \multirow{2}{*}{\shortstack[l]{Change point\\ detection}}&\multirow{2}{*}{PELT-algorithm}&\multirow{2}{*}{$\beta_{primary} = 3$}&\multirow{2}{*}{$\beta_{secondary} = 1.5$}\\&&&\\\hline
          \multirow{2}{*}{\shortstack[l]{Cause-effect\\ analysis}}&\multirow{2}{*}{Granger-causality}&\multicolumn{2}{c|}{\multirow{2}{*}{$\text{p-value}{=}1\times 10^{-12}$}}\\
          &&\multicolumn{2}{c|}{}\\\hline
    \end{tabular}
    }
    \label{tab:par_synth}
\end{table}
}

For change point detection we retrieve a change point in the primary control-flow perspective at day 133 and in the secondary data perspective at day 132. The lag between this drift is $k{=}1$. We, therefore, use a lag of 1 when testing for a cause-effect relationship between the two perspectives.

\begin{figure}[t]
\centering
    \includegraphics[width=.99\textwidth]{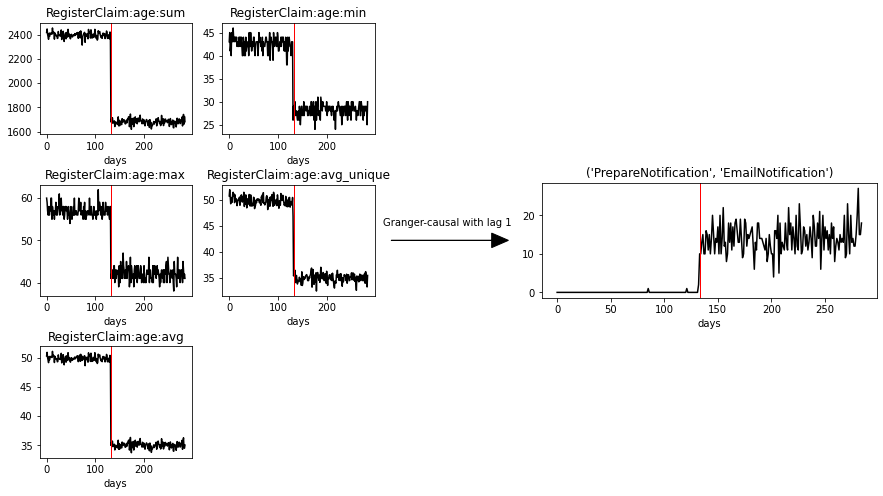}\hfill 
    \caption{5 Granger-causal feature pairs for the cause-effect relationship between data and control-flow perspective. A drift in the age of the customers is responsible for an increase in email notifications and a decrease in other notifications.}\label{fig:results_synthetic}
\end{figure}

With a p-value of $10^{-12}$, which is especially low due to the artificial setting, we retrieve 25 feature pairs that are Granger-causal with lag $k{=}1$. All involved features of the primary control-flow perspective concern the frequency of directly follows relationships between one of the notification activities and either a preceding or succeeding activity. The features of the secondary perspective all describe the distribution of age, i.e., the sum, average, minimum, maximum and average of the set of values. We, therefore, limit our output to only 5 of the feature pairs, which are depicted in \autoref{fig:results_synthetic}. The depicted features propose that a decrease in the age of customers led to an increase in the prevalence of the email notification activity one day later which is exactly the cause-effect relationship we artificially introduced.
Our framework has correctly detected and explained the concept drift with the underlying cause-effect relationship.

\subsection{Case Study}
We also evaluated our framework using data from the Business Process Intelligence (BPI) Challenge 2017 \cite{vanDongen2017}. The event log considered belongs to a loan application process through an online system.  A customer can submit loan applications to the financial institute and may receive an offer from the financial institute afterwards.
The parameters for applying our framework are depicted in \autoref{tab:par_real}.
We  analyze the performance perspective, i.e., the service times, for concept drifts. We search for root causes in the resource perspective, i.e., the workload, as it has shown to often have a significant impact on the service times \cite{analyzingresourceBehaviourUsingProcessMining}.

\begin{figure}[t]
\centering
    \includegraphics[width=.99\textwidth,valign=t]{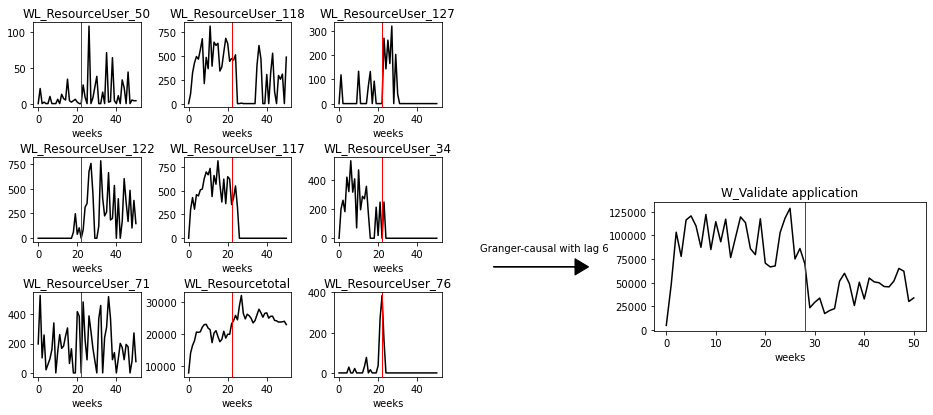}\hfill
    \caption{Granger-causal feature pairs of the cause-effect relationship between resource and performance perspective. The only concept drift for the duration perspective of these features can be observed in \textit{W\_Validate application}. An increase of the workload is Granger-causal to the reduction of the service times.}\label{fig:results_bpi2017}
\end{figure}

For the primary, performance perspective we retrieve a change point in Week 28. For the secondary, workload perspective we retrieve a change point at Week 22. The lag for cause-effect analysis is therefore $k{=}6$.
The cause-effect analysis with a p-value of $0.015$ yields 23 Granger-causal feature pairs with a lag of 6 weeks. Four different primary features are contained in these feature pairs. Since three of them do not exhibit a concept drift around week 28, we drop the corresponding pairs for further analysis. The remaining pairs are depicted in \autoref{fig:results_bpi2017}. The average duration of \textit{W\_Validate application} shows a significant decrease for week 28. We further analyze the resource workloads that are Granger-causal to this feature. Most of the resources do not work continuously over the span of the event log. We can see increases and peaks in the workload for some resource. When looking at the total workload of all resources, which is among the Granger-causal features, we can see a significant increase. The detected cause-effect relationship, therefore, states that \textbf{an increase in the workload of resources led to a decrease in the service times for \textit{W\_Validate application}}.

One submission paper for the BPI Challenge \cite{bergerbpi2017},  amongst other things, investigates different KPIs of the process over time. This paper also found a significant decrease in the manual validation time, i.e., the service times for validating an application, and an increase in case numbers. Due to the absence or the lack of visibility of other factors such as additional training, change of staff, etc., the paper suggests that the decrease in service times is a reaction to cope with the increased workload.

\setlength{\tabcolsep}{0.35em} 
{\renewcommand{\arraystretch}{1.2}
\begin{table}[t]
    \centering
    \caption{Parameter choices for running the framework on the BPI 2017 log.}
    \resizebox{0.8\textwidth}{!}{
    \begin{tabular}{|l||l|l|l|}\hline
    \multicolumn{4}{|c|}{Experimental Setup}\\\hline
         \multirow{3}{*}{\shortstack[l]{Time series\\ construction}}&Primary perspective& Performance& Service times \cite{analyzingresourceBehaviourUsingProcessMining}\\\cline{2-4}
         &Secondary perspective& Resource & Workload \cite{analyzingresourceBehaviourUsingProcessMining}\\\cline{2-4}
         &Time interval duration & \multicolumn{2}{c|}{1 week}\\\hline
          \multirow{2}{*}{\shortstack[l]{Change point\\ detection}}&\multirow{2}{*}{PELT-algorithm}&\multirow{2}{*}{$\beta_{primary} = 3$}&\multirow{2}{*}{$\beta_{secondary} = 1.5$}\\&&&\\\hline
          \multirow{2}{*}{\shortstack[l]{Cause-effect\\ analysis}}&\multirow{2}{*}{Granger-causality}&\multicolumn{2}{c|}{\multirow{2}{*}{$\text{p-value}{=}0.015$}}\\
          &&\multicolumn{2}{c|}{}\\\hline
    \end{tabular}
    }
    \label{tab:par_real}
\end{table}
}

\subsection{Comparison}
Our proposed framework for explainable concept drift detection touches two areas of research: Concept drift detection and cause-effect analysis. We, therefore, compare the results for the synthetic event log and the BPI 2017 log with the results from state-of-the-art methods from both of these areas. For concept drift detection we compare the results with the visual analytics approach of Yeshchenko et al. \cite{ProcessDriftVisualAnalytics} and ProDrift by Maaradji et al. \cite{CPDHypothesisTestingAdaptiveWindow}  as both approaches have shown outstanding results in concept drift detection. For cause-effect analysis we compare our results with the findings of Pourbafrani et al. \cite{PMSD} as they are searching for relations between different process parameters on a system-wide level.

\begin{table}[t]
    \centering
    \caption{Comparison of results for the concept drift detection}
    \resizebox{0.8\textwidth}{!}{
    \begin{tabular}{|l|l|c|c|}
    \hline
   
     & Our Approach & Visual Analytics \cite{ProcessDriftVisualAnalytics} & ProDrift \cite{suddengradualdriftostovar}\\ \hline
      Synthetic Log   &  Control-flow drift, day 133& \checkmark & \checkmark \\\hline
      BPI 2017   & Performance drift, week 28 & $\times$ & $\times$ \\\hline 
    \end{tabular}
    }
    \label{tab:cpd_results_comparison}
\end{table}

\begin{table}[t]
    \centering
    \caption{Comparison of results for cause-effect analysis}
    \resizebox{0.65\textwidth}{!}{
    \begin{tabular}{|l|l|c|}
    \hline
    
     & Our Approach & PMSD \cite{PMSD}\\ \hline
      Synthetic Log   & Data (age) $\rightarrow$ control-flow & $\times$  \\\hline
      BPI 2017   & Resource (workload) $\rightarrow$ performance & $\checkmark$ \\\hline 
    \end{tabular}
    }
    \label{tab:cae_results_compare}
\end{table}

\autoref{tab:cpd_results_comparison} depicts the comparison between the detected concept drifts for the synthetic and the real-life event log. The control-flow drift in the synthetic log is detected by both approaches. As ProDrift relies on completed traces, the drift is detected approximately 15 days later compared to our approach. Both approaches very clearly show the existence of a sudden drift through means of their visualization. As both approaches do only focus on control-flow drifts they can not be used to compare results on the detected performance drift for the BPI 2017 log.
\autoref{tab:cae_results_compare} depicts the results retrieved from PMSD compared to our approach. As PMSD does not model the data perspective, we can not use it to detect the cause-effect in the synthetic log. For the BPI 2017 log we apply the PMSD framework and compute a system dynamics log. This log contains, among others, the arrival rate and the service times of the process. We apply the relation detection of PMSD with a lag of 6 weeks. The results show a negative correlation between the lagged arrival rate and the service times. This corresponds to our detected cause-effect as the higher influx of cases determined a decrease in service times.

We verified our findings by applying state-of-the-art methods from both concept drift detection and cause-effect analysis. If the corresponding perspective can be modeled, we are able to verify our findings with these approaches. These are promising results as they show the power of incorporating more perspectives into concept drift detection and using these to find possible cause-effects of concept drifts.

\section{Conclusion}
\label{sec:conclusion}
In this paper, we combine concept drift detection and cause-effect analysis to create a framework for explainable concept drift detection. We define a primary perspective where concept drifts should be detected and a secondary perspective with which these concept drifts are explained. By applying a cause-effect analysis to the features of both perspectives, we identify feature pairs that can be used to explain the concept drift. We verified our approach using a synthetically generated event log. We furthermore analyzed the event log of the BPI Challenge 2017 and were able to explain a concept drift in the performance with an increase in the resources' workload. These first experiments have shown a great potential for explaining concept drifts. 
\paragraph{Future Work}
To further improve our conceptual framework the following steps can be taken. First of all, we want to plug different change point detection algorithms and cause-effect analysis tools to detect other types of drifts and, e.g., non-linear relationships. Furthermore, spurious elements and rare signals produce spikes in a signal that can be misleading to cause-effect analysis techniques. We want to investigate whether the general application of noise filtering on the time series is beneficial. Another interesting point for an extension of the framework are non pairwise dependencies. A concept drift could, e.g., be caused by two different concept drift in two other perspective and not by only one of them. 

\bibliographystyle{splncs04}
\bibliography{bibliography}

\end{document}